\documentclass{article}
\usepackage[utf8]{inputenc}
\usepackage[a4paper,margin=2.5cm]{geometry}

\usepackage[round ]{natbib}
\usepackage{multirow}
\usepackage{graphicx}

\title{Code Word Detection in Fraud Investigations using a Deep-Learning Approach}
\author{Y. van der Zee, J.C. Scholtes, M. Westerhoud and J. Rossi}
\date{March 2021}

\begin{document}

\maketitle

\begin{abstract}
In modern litigation, fraud investigators often face an overwhelming number of documents that must be reviewed throughout a matter. In the majority of legal cases, fraud investigators do not know beforehand, exactly what they are looking for, nor where to find it. In addition, fraudsters may use deception to hide their behaviour and intentions by using code words. Effectively, this means fraud investigators are looking for a needle in the haystack without knowing what the needle looks like.

As part of a larger research program, we use a framework to expedite the investigation process applying text-mining and machine learning techniques. We structure this framework using three well-known methods in fraud investigations: (i) the fraud triangle (ii) the golden ("W") investigation questions, and (iii) the analysis of competing hypotheses. With this framework, it is possible to automatically organize investigative data, so it is easier for investigators to find answers to typical investigative questions.

In this research, we focus on one of the components of this framework: the identification of the usage of code words by fraudsters. Here for, a novel (annotated) synthetic data set is created containing such code words, hidden in normal email communication. Subsequently, a range of machine learning techniques are employed to detect such code words. We show that the state-of-the-art BERT model significantly outperforms other methods on this task. With this result, we demonstrate that deep neural language models can reliably (F1 score of 0.9) be applied in fraud investigations for the detection of code words.

\end{abstract}

\section{Background}

In fraud investigations, investigators have to deal with ever increasing mass of unstructured data. These data can be a valuable source of information and sometimes even direct evidence in relation to the matter that is investigated. More often than not, these investigations are supported by eDiscovery tools. The development of AI-techniques to enhance these tools is primarily aimed at isolated topics, such as sentiment and emotion analysis, assisted review (searching using machine learning), or Named Entity Recognition (NER). Although these techniques as such are very promising, there is no direct link to the way an investigator approaches a fraud investigation. 

In this paper we present a model that grants AI-techniques a logical role in a fraud investigation. For this, let us first have a look how a typical investigator approaches an investigation. This can be done by examining three building blocks that provide a basis where we can ‘plug in’ an AI-technique and use the outcome as a diagnostic variable in the investigated case. These building blocks are: (i) the Fraud Triangle \citep{cressey2-1951} \citep{Kassem2012}, (ii) the six (6) ‘Golden’ investigation questions and (iii) the Theory of the Analysis of Competing Hypotheses \citep{heuer1999psychology}. With these building blocks we can deconstruct a (partial) investigation question into a number of tasks that can each be executed by a specific search, text-mining or machine-learning algorithm. Let us first explain these three components individually, and then explain how we combine them.

\subsection{Fraud Triangle}

\begin{figure}[ht]
    \centering
    \includegraphics[width=\textwidth]{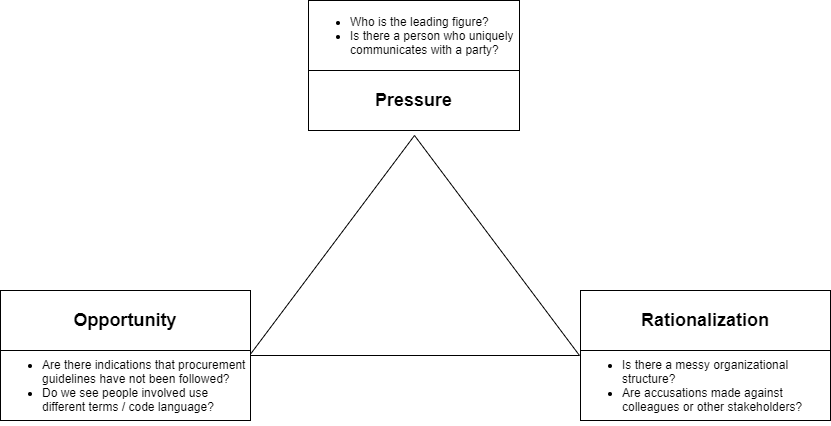}
    \caption{The Fraud Triangle with relevant investigation questions}
    \label{fig:fraud_triangle}
\end{figure}

A widely used method to model organizational fraud risk is the fraud triangle, visualized in figure \ref{fig:fraud_triangle}. Just as fire requires fuel, oxygen and a spark, in the case of a fraud there are also three ingredients are essential: The perpetrator must have a \textbf{motive} to commit fraud, the situation must provide an \textbf{opportunity}, and the fraudster must find a way for herself to \textbf{rationalize} her dishonesty. Motives can vary from perverse financial incentives to personal problems, such as financial need or addiction. All these can be referred to as \textbf{pressure}. The opportunity is often related to the control environment of the victim organization: weak controls and tone at the top. And finally the rationalization relates to the perceived relation between the fraudster and his environment. This relation provides the internal justification of a fraud: ‘I was mistreated’, ‘everybody does it’, etc.

\subsection{Six (6) Golden Investigation Questions}
Usually the fraud triangle is used as a risk tool. But, we can also use the model as part of our investigation framework. To do this, we propose a relationship between the three (3) edges of the fraud triangle and the six golden questions that lie at the basis of almost every fraud investigation: who, why, what, how, when and where. Answering these question will almost automatically lead to the construction of a possible fraud scenario and fill the elements of an evidence matrix. If one is in need to know what the motives of a fraudster are, one needs to know \textbf{who} did it and \textbf{why}. If one needs to know about possible fraud opportunities, questions about the \textbf{what} and \textbf{how} need to be answered. And finally, for the rationalization component of the fraud triangle, situational variables are important, in particular: \textbf{where} and \textbf{when}.

\begin{table}[h]
\centering
\begin{tabular}{|l|l|}
\hline
\multirow{2}{*}{Motive}          & Who did it?          \\ \cline{2-2} 
                                 & Why did she do it?   \\ \hline
\multirow{2}{*}{Opportunity}     & What happened?       \\ \cline{2-2} 
                                 & How did he do it     \\ \hline
\multirow{2}{*}{Rationalisation} & Where did it happen? \\ \cline{2-2} 
                                 & When did it happen?  \\ \hline
\end{tabular}
\caption{Combining the Fraud Triangle (left side of table) with the Golden Investigation Questions (right side of table)}
\end{table}

Answers to (variations of) these questions produce evidence items that can populate elements of the evidence matrix.

\subsection{Analysis of Competing Hypotheses (ACH)}
For each type of crime, a so-called evidence matrix can be constructed holding key items to be proved. For instance, in case of a murder one needs a victim, a murder weapon, a motive, a crime scene, intent, etc. These items relate to the above mentioned Golden Investigation Questions. Instead of using a simple numeration of such items, we can use a more advanced model of an evidence matrix as developed in the 1970’s by Richard Heuer \citep{heuer1999psychology}. This methodology was named: Analysis of Competing Hypotheses (ACH). It is based on the evaluation of various competing hypotheses, given a set of information items (i.e. evidence). This involves the following step-by-step approach:

\vspace{5mm} 
\textbf{Step-by-Step Outline of Analysis of Competing Hypotheses:}

\begin{itemize}
    \item Identify the possible hypotheses to be considered. Use a group of analysts with different perspectives to brainstorm the possibilities.
    \item Make a list of significant evidence and arguments for and against each hypothesis.
    \item Prepare a matrix with hypotheses across the top and evidence down the side. Analyze the "diagnosticity" of the evidence and arguments--that is, identify which items are most helpful in judging the relative likelihood of the hypotheses.
    \item Refine the matrix. Reconsider the hypotheses and delete evidence and arguments that have no diagnostic value.
    \item Draw tentative conclusions about the relative likelihood of each hypothesis. Proceed by trying to disprove the hypotheses rather than prove them.
    \item Analyze how sensitive your conclusion is to a few critical items of evidence. Consider the consequences for your analysis if that evidence were wrong, misleading, or subject to a different interpretation.
    \item Report conclusions. Discuss the relative likelihood of all the hypotheses, not just the most likely one.
    \item Identify milestones for future observation that may indicate events are taking a different course than expected.

\end{itemize}

\begin{figure}[ht]
    \centering
    \includegraphics[width=\textwidth]{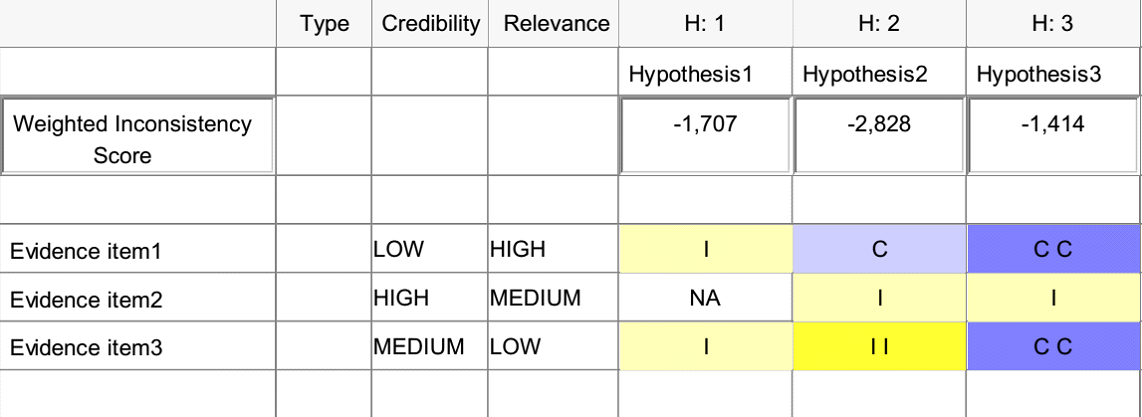}
    \caption{Visualization of the ACH Matrix. I	Inconsistent, II	Strongly inconsistent, C	Consistent, CC	Strongly consistent, NA	Not applicable}
    \label{fig:my_label}
\end{figure}

The ‘weighted inconsistency score’ provides a measure for the plausibility of a specific hypothesis, given a set of evidence items in terms of \textit{credibility} and \textit{relevance}. Lower values of the scores, correspond with a lower plausibility of the hypothesis. The numerical values are determined based on a simple lookup table. These initial values do not represent probabilities, but they can be normalized towards a [0-1] range, giving a normalized confidence score. Combining confidence scores can be done by multiplication. There are obvious issues with this approach, as the use multiplication in the calculations presumes complete independence of the underlying hypothesis, which is off course not always the case. In addition, the values are manually assigned, which leads to bias risks. But for now, this is what is used.\footnote{See: http://www.pherson.org/PDFFiles/ACHTechnicalDescription.pdf for a technical discussion on the use of 'weighted inconsistency scores'}  

Now we have a conceptual model we can systematically inject the results of a various set of AI-methods into. For example in a case of the investigation of a possible purchasing scheme. Typical for this scheme is the incidence of collusion between perpetrators. 

Several of the evidence components listed above, can be filled automatically with possible candidates using basic text mining techniques. For example: Named Entity Extraction in combination with basic linguistic contextual analysis can provide candidates for the Who, Where, and When questions. Sentiment and emotion mining can identify the textual sections containing providing valuable insights into Why something is done and Who is driving the actions. 
On-going research focuses on the creation a further mapping between fraud types being investigated, related questions to populate the scenario’s and AI-methods that can be applied to generate evidence items in answering these questions. But initial results are promising as is described in \citep{scholtes2020} and \citep{scholtesherik2021}.

Often these accomplices use code words in their communications. Finding these, is actually the beginning of a fraud investigation and an essential first step to truly understand the meaning of extracted information in the context of a crime investigation. For example, if the suspects communicate about "players" instead of "employees that can be bribed" and about "how much coffee did you give him" referring to monetary compensation. It is essential to understand such code words before one is truly able to fill the ACH-matrix with possible candidates.

Therefore, we are interested in the use of text-mining and machine learning techniques for the automatic and domain-independent detection of code words.  

This element of the scenario can be constructed as follows when added to the ACH-matrix:

\begin{table}[ht]
\begin{tabular}{|l|l|}
\hline
& \begin{tabular}[c]{@{}l@{}}\textbf{Question:} \\ Who is likely to be involved in the scheme?\end{tabular}        \\ \hline
\begin{tabular}[c]{@{}l@{}}\textbf{AI-method:} \\ code word detection\end{tabular} & \begin{tabular}[c]{@{}l@{}}\textbf{Result:} \\ Use of code words between employee A and employee B and supplier C\end{tabular} \\ \hline
\end{tabular}
\end{table}

Entering this result as an evidence item in an ACH-matrix would look like this:

\begin{table}[ht]
\begin{tabular}{|l|l|l|l|l|}
\hline
                                                                               & Credibility & Relevance & \begin{tabular}[c]{@{}l@{}}H1:\\ A is acting alone\end{tabular} & \begin{tabular}[c]{@{}l@{}}H2:\\ A and B are colluding\end{tabular} \\ \hline
\begin{tabular}[c]{@{}l@{}}Use of code words\\  between A, B and C\end{tabular} & High        & Medium    & I                                                               & CC                                                                  \\ \hline
\end{tabular}
\end{table}

An additional goal of this research is to avoid the need for specific domain or world knowledge when detecting code words: we wish to recognize any type of code words in any kind of context. We do understand that there will be a language dependency (is communication for instance in English or in German?), but by using general language models, we wish to avoid the need for specific domain-dependent language models. 

\section{Introduction}

The ability to model the context of a text is vital in fraud investigations, especially for code-word detection. The ability to properly model this context has greatly advanced in recent years due to the successful advances using deep-learning algorithms for highly context sensitive Natural Language Processing (NLP) tasks such as machine translation, human-machine dialogues or co-reference and pronoun resolution. For this reason, we are interested to investigate the application of these techniques towards the detection of code word detection by bench marking their performance against existing baseline models. 

The above mentioned progress mainly originates from the development of the so-called Transformer architecture. Transformer models are based on large pre-trained networks that already embed significant linguistic knowledge and which can be fine-tuned on specific tasks requiring a relatively small amount of additional training. Let us explain how this is done:

\begin{itemize}
\item A fundamental benefit of the transformer architecture is the ability to perform Transfer Learning. Traditionally, deep learning models require a large amount of task-specific training data in order to achieve a desirable performance (billions of data points required to fine tune hundreds-of-millions neural interconnections). However, for most tasks, we do not have the amount of labeled training data required to train these networks. By pre-training with large sets of natural text, the model learns a significant amount of task-invariant information on how language is constructed. With all this information already contained in these models, we can focus our training process on learning the patterns that are specific for the task at hand. We will still require more data points than required in most statistical models (typically 50-100k based on our experience in earlier NLP Deep-Learning projects), but not as much as the billions required, should we start the training of the deep-learning models from scratch. 
\item Transformers are able to model a wide scope of linguistic context, both depending on previous words, but also on (expected) future words. They are, so to say, more context sensitive than models that can only use past-context into consideration. In addition, this context is included in the embedding vectors, which allows for a richer representation and more complex linguistic tasks.
\end{itemize}

By the end of 2018, researchers at Google AI Language released a new model named the Bidirectional Encoder Representations from Transformers (BERT) \citep{Devlin2019}. In the subsequent months, BERT models achieved impressive benchmark performance on most downstream NLP tasks. Currently, BERT is considered to be the state-of-the-art language representation model. For this reason, we aim to apply BERT to the task of code word detection. 

\section{Related Work}

Published research on code-word detection in criminal investigations is sparse. \citet{Jabbari2008} proposes a distributional model of word usage and word meaning, derived purely from a corpus of the text. This model is applied to determine whether certain words are used \textit{out of context}. To be more specific, the authors generate four different test sets, choosing one specific word for each test set and selecting 500 sentences containing the chosen word. These sentences are considered to represent the normal meaning of the word. Next, the authors substitute a second, unrelated word for the chosen word  in 500 other sentences to artificially create examples of word obscurities. The authors show that a certain understanding of the context is captured by the distributions, and that this understanding is valuable in detecting obfuscated words. No overall quantitative results are reported.  

In the work of \citet{FongSzeWangandRoussinovDmitriandSkillicorn2008}, the authors follow a similar approach by randomly sampling sentences from the Enron data set and replacing the first noun in each sentence with a different noun. On a test set of a few hundred samples, the authors achieve a detection ratio of 83\% and 90\% on two other synthetic data sets.

Two more recent works \citep{Magu2017, Magu2018}, focus on the detection of euphemistic hate speech on social media platforms. Euphemistic hate speech is distinct from other forms of implicit hate speech because in reality they are often direct poisonous attacks as opposed to veiled or context-dependent attacks. They are implicit because they use clever word substitutions in the language to avoid detection. The authors use word embeddings and network analysis to identify these word euphemisms. Here too, no overall quantitative results are reported. 

To our knowledge, no work has been done on detecting code words using large pre-trained deep-learning models such as BERT.

\section{Method}


We follow a similar methodology as \citet{FongSzeWangandRoussinovDmitriandSkillicorn2008}. However, we make the necessary adjustments to generate significantly more samples that we can use as training data compared to  their work. As explained earlier, we anticipate to need for about 50k training data points.  As a starting point, we randomly sample the content of emails from the ENRON data set. Here, we limit our attention to the substitution of nouns since they are the more likely candidates to be substituted by code words. This then results in a sentence which contains one word that is out of context. Consider the example: it is not difficult to imagine a scenario in which colleagues use a nickname or code word for their office, such as "the rock". This will result in the following substitutions in their email communication: \textit{I will be out of the office on Friday}. Here, we take the noun \textit{office} and replace it with a different noun, for example, \textit{rock}. The result is \textit{I will be out of the rock on Friday}, a sentence in which the word rock is an outlier concerning the rest of the words. These are precisely the kind of examples our models are supposed to identify.

We use the methodology described above to generate a data set that can be used to train and evaluate different machine learning techniques. A second synthetic data set is used to investigate the extent to which the trained models can be used in practice, with the aim of testing the approaches in a more realistic scenario.  In this scenario, three types of drugs are mentioned, namely cocaine, marijuana and heroin. Individuals communicating have the goal to hide the fact that they are talking about an illegal drug. Therefore, they use code words to describe the drugs. We believe this scenario is more realistic than the first synthetic data set because in this scenario individuals specifically avoid certain terminology in their communication.
To create realistic code words, we use a list of Drug Slang words provided by the U.S. Drug Enforcement Administration (DEA) \footnote{https://ndews.umd.edu/sites/ndews.umd.edu/files/dea-drug-slang-code-words-may2017.pdf}. As an example, imagine the following text message; \textit{I'm about to buy some cocaine for our party tonight; see you there}. We replace the instance of the word cocaine with snow, one of the code words for cocaine, resulting in the following sentence; \textit{I'm about to buy some snow for our party tonight; see you there}. Ideally, our trained models will identify this sentence as an example of code word usage, since the word snow does not belong in this context.

\subsection{Synthetic Data sets}

\subsubsection{ENRON Code word data set}

The ENRON data set\footnote{https://www.cs.cmu.edu/~./enron/} was made public as a result of the prosecution of ENRON personnel. It contains about half a million emails to and from ENRON employees over three and a half years. The authors of the emails never expected them to be made public, so it is an excellent example of informal writing. The data set contains emails from a large number of authors, from many backgrounds. As such, it is a good surrogate for the type of messages that could be intercepted in a fraud investigation. 

We randomly sample body messages of emails from the data set and evaluate each sentence individually. We only consider sentences containing between five and twenty words and apply part-of-speech tagging to verify if the sentence contains a noun. If the sentence is of the desired length and contains a noun, it will be added to our pool of candidate sentences. We keep sampling until the candidate pool consists of 60,000 sentences. For half of the candidate sentences, we replace the first noun with a noun taken from the BNC noun list\footnote{
The British National Corpus (BNC) was originally created by Oxford University press in the 1980s - early 1990s, and it contains 100 million words of text texts from a wide range of genres (e.g. spoken, fiction, magazines, newspapers, and academic). https://www.word frequency.info/compare\_bnc.asp}. To ensure that the models do not over-fit on this limited list of nouns, we split the BNC noun list into train, validation, and test subsets and sample a substitute word from the appropriate sub-list. We complete the train, validation, and test set with the other half of the candidate sentences, in which no words have been replaced. These sentences will serve as negative samples. The training and validation sets consist of 48,000 and 6,000 samples, respectively, with equal samples from each class. The test set consists of 6,000 samples, but here we introduce a class of imbalance to make the setting more like a typical eDiscovery data set. The set contains 400 positive samples, which means that only 5\% of the samples contain a code word.

\subsubsection{Reddit Drugs Data set}

To create the Reddit Drugs Data set, we extract user-submitted Reddit comments from the subreddit \textit{/r/worldnews}. The PushShift API\footnote{https://github.com/pushshift/api} is utilized to gather comments before April 2020. As with the selection of candidate sentences for the ENRON Code word data set, we select candidate sentences of between five and twenty words. To make sure the sentence originates from the English language, we use the Spacy language detector\footnote{https://spacy.io/universe/project/spacy-langdetect} to detect the language of the comment. The first 600 sentences that meet these criteria are added to our data set and will act as our negative samples.

The same process follows, only this time, we use PushShift's search function to search for comments that mention three pre-selected drugs. The chosen drugs are Cocaine, Marijuana, and Heroin. The same pre-selection is applied to each sentence of each comment, and if the sentence contains one of the drugs, the sentence is added to the candidate list. Each sentence is then manipulated by replacing each mention of all three drugs with a code word. The code words chosen for each drug are shown in Table \ref{tbl:code words_reddit}. Each modified sentence is added to the data set as a positive example. The result is a balanced data set of 1200 samples with 600 samples of each class.

\begin{table}[h]
\centering
\begin{tabular}{|l|l|}
\hline
\textbf{Meaning} & \textbf{Code word in text} \\ \hline
Cocaine             & Line                \\ \hline
Marijuana              & Bush                  \\ \hline
Heroin               & Pure              \\ \hline
\end{tabular}
\caption{The selected code words for the Reddit Drugs data set.}
\label{tbl:code words_reddit}
\end{table}

\subsection{Models}

In this research, we will benchmark four commonly used NLP models on the code word detection task. The four selected methods are Bag of Words (BoW), TF-IDF, BiLSTM initialized with GloVe embeddings and BERT. With these selected models, we transition from shallow lexical approaches to dense contextual text representations. BERT was mainly chosen over other contextual methods like ELMo \citep{peters2018deep} as it is considered state-of-the-art on most downstream NLP tasks. In addition, we prefer BERT over ELMo because the use of sub words allows the former to better deal with out-of-vocabulary words. We benchmark each of the models' performance to determine the importance of contextual representations on the task of code word detection. 

Both the BoW and TF-IDF method are used in combination with a Logistic Regression classifier. Both methods consider unigrams, bigrams and trigrams and ignore terms that have a frequency lower than three. The deep learning models are implemented using the PyTorch framework \citep{paszke2019pytorch}. The BiLSTM model uses an embeddings layer that is initialized with pre-trained GloVe embeddings. The model is trained on binary cross-entropy loss until convergence using the Adam optimizer \citep{Kingma2015}. The BERT implementation uses the \textit{bert-base-uncased} configuration from the hugging face transformers package\footnote{https://github.com/huggingface/transformers}. This specific model was preferred over other models due to its size. All the encoder layers are updated during training. The fine-tuning of BERT is done using the standard settings for fine-tuning, namely training for ten epochs and updating the model parameters using the Adam optimizer with a learning rate of $2e-5$ and $\epsilon = 1e-8$.

\subsection{Evaluation Metrics}

For this task, the metrics used are the standard Accuracy, Precision, Recall, and F1 scores. In addition, we assess the results on the unbalanced data set using macro-averages. Macro-averaging computes the metrics independently for each class and then take the average, thus treating all classes equally. We select macro-averaging over micro-average, where the contributions of all classes are aggregated to compute the average metric, as we are primarily interested in the performance of the models on the code word class. Lastly, with this interest in mind, we evaluate the models on the class-specific precision and recall of the code word class (C1 Precision \& C1 Recall). Our primary interest is to achieve high recall on this class, as we want to recover all samples that may contain a code word. Beyond that, it is also essential to achieve reasonable precision on the positive class, because otherwise the user will have to navigate a sea sheet positively.

\section{Results}

Table \ref{tbl:code words-rq1.1} shows the performance of the selected models on the validation set. As discussed, this set is generated using the same methodology as the training set, only using a different set of nouns to replace with. We observe that both lexical approaches perform significantly better than a naive random classifier. This is a somewhat surprising result as one would expect that these models do not model the essential context needed for this task. It seems that for a number of examples, the lexical approaches provide sufficient information to differentiate between the use of code words for some of the samples. 

We see the first significant performance increase with the Bi-LSTM model. This performance increase can be attributed to the use of pre-trained word embeddings. As discussed, word embeddings are learned by predicting words given the surrounding context. This task is related to our task in a big way, as we want to find words that do not belong in a given context. Since word embeddings capture information about which words are being used together, they help us find better examples of sentences using one word from the context. 

\begin{table}[h]
\centering
\begin{tabular}{|l|l|l|l|l|l|l|}
\hline
  Model      & Accuracy & Precision & Recall & F1-Score & C1 Precision & C1 Recall \\ \hline
Random  & 0.50      & 0.50        &  0.50  &   0.50  & 0.50 &  0.50   \\ \hline

BoW     &  0.63        &  0.64         & 0.63      &     0.62 & 0.67 &  0.50    \\ \hline
TF-IDF  &  0.63       &   0.63     &  0.62   &    0.62   &  0.52 & 0.58   \\ \hline
Bi-LSTM &  0.80        &   0.80        &   0.80     &   0.80   & 0.83 & 0.76     \\ \hline
BERT    & \textbf{0.90}      & \textbf{0.90}        &    \textbf{0.90}    & \textbf{0.90}  & \textbf{0.95} & \textbf{0.84}    \\ \hline
\end{tabular}
\caption{Results on the ENRON Codeword Validation Set.}
\label{tbl:code words-rq1.1}
\end{table}

We observe a significant performance increase from the pre-trained BERT model. As discussed in the background section, BERT has demonstrated that it performs better on most downstream NLP tasks than the previous state-of-the-art models, and for our task, this is also the case. This significant increase can be attributed to two aspects of the model. First, deep neural language models are larger than the other models and can learn more parameters, which means more stored information. Secondly, because of how BERT is trained, the model can capture the context of the sentence very well. Because this context information is vital to our task, the BERT model is best suited to this task, as demonstrated by its performance on the validation set.

\begin{table}[h]
\centering
\begin{tabular}{|l|l|l|l|l|l|}
\hline
    Model    & Macro Precision  & Macro Recall & C1 Precision & C1 Recall & C1 F1-Score \\ \hline
Random     &  0.50    &  0.50      &  0.05     & 0.50 &  0.09      \\ \hline
BoW     & 0.54       &  0.64       &  0.12      & 0.49   &   0.19    \\ \hline
TF-IDF  &  0.53 &   0.62     &  0.09       & 0.50     & 0.15     \\ \hline
Bi-LSTM &  0.60        &   0.82        &  0.22      & 0.79 &  0.34        \\ \hline
BERT    & \textbf{0.79}    & \textbf{0.90}   & \textbf{0.59}   &  \textbf{0.84}  & \textbf{0.69}  \\ \hline
\end{tabular}
\caption{Results on the ENRON Codeword Test Set.}
\label{tbl:code words-rq1.2}
\end{table}

We observe similar performances on the imbalanced test set, as shown in table \ref{tbl:code words-rq1.2}. The models follow a similar trajectory in terms of macro precision and recall performance. However, both count-based approaches achieve a precision of around 0.1 on the code word class and a random performance in terms of recall on samples from that class. This performance is to be expected since we work with an unbalanced dataset. However, within eDiscovery, we virtually always work with unbalanced datasets, which makes it extremely important that a method also works for unbalanced data. We prefer a high recall on the code word class since no samples must be missed during a review. Of course, it is also essential that reasonable precision is achieved. Otherwise, the user will mainly see false positives. We observe a significant performance increase from the BERT model on this unbalanced dataset, especially in terms of recall on the code word class. From these results, we can conclude that the BERT model significantly outperforms the other models on all selected metrics.

\section{Conclusions}

To conclude, we will cover the main findings of our study. We have observed how BERT is able to capture richer contextual representations than the other bench-marked models. This results in higher performance on the code word detection task. But more importantly, this new approach creates the  possibility to detect patterns that were previously not automatically detected (highest recall). Detecting code words is one example of this, but as discussed earlier, there are many other examples of these fraudulent patterns within fraud investigations. The method is domain independent, so any type of code words will be recognized. In fact, the BERT-based method will pick up any kind of unexpected language use. 

There is the more philosophical questions on how well the BERT model really \textit{understands} the code words. On the one hand, we cannot under-estimate BERT's ability to memorize due to the enormous amount of connections that allow it to memorize word patterns. \textit{Clever Hans} (in German, \textit{der Kluge Hans}) was a horse that was supposed to be able to do lots of difficult mathematical sums and solve complicated problems. Later, it was discovered that the horse was giving the right answers by watching the reactions of the people who were watching him. By observing reactions of the human, the horse “answered” correctly. But, in reality the horse only responded with memorized behavior. Is BERT just a clever Hans by “memorizing” word patterns without really understanding this? On the other hand, we can ask ourselves how much of our human language skills can be attributed to memorization \citep{DBLP:journals/corr/abs-1907-07355} \citep{heinzerling2019cleverhans}. 

In conclusion: from a pragmatic point of view we can observe that by teaching a valid language model for a certain language, BERT can successfully learn to recognize deviating language patterns given this language model. In the context of this research, those are code words used by fraudsters. But it is fair to say, that other patterns are also picked up, for instance non-English language, misspellings, and scanning-related Optical Character Recognition (OCR) errors. So, the model we developed appears to be more a general model to pick up \textit{unexpected} language patterns, instead of a model that explicitly \textit{understands} and \textit{detects} code words. But for the \textit{application} of code word detection, it can be deployed successfully.

\section{Discussion}

While these initial results are promising for the application of the detection of code words, much remains to be done to successfully use machine-learning techniques in fraud investigations. One of the biggest research challenges in the AI research community is the challenge of explainable AI. Due to the sheer size of these neural networks, it is extremely difficult to figure out why the model makes a certain classification and what exactly this classification is based on. Therefore, in this research we choose to use machine learning techniques to support fraud investigation rather than replace the human investigator. As a result, the benefit to the fraud investigation is twofold. By setting up the framework, the fraud investigator brings structure to the search space, and with the employment of machine learning techniques, this search space is reduced and made insightful. At the same time the framework offers cohesion to the collection, classification and weighting of evidence that is collected via AI-methods.

Further research will be done to identify more relevant evidence items which have a discriminative relation to a fraud scenario and which can be obtained by an appropriate AI-method. In our proposed model these evidence items all are formatted as an answer to one or more variants of the six golden investigation questions. With an adequate amount of these 'triples' (scenario-evidence-AI-method) we expect that many investigations can benefit significantly in terms of efficiency and quality. Another topic for further development is the automation of applying weights in terms of relevance and credibility to the output of the AI-method and subsequently inserting consistency values into the ACH-matrix. 

But, by being able to detect code words, a great first step is made into the proper interpretation of potential relevant patterns that can be used to fill other components of the ACH-matrix. 

\section{Acknowledgements}

The authors are grateful for  the extensive support obtained for this research from ZyLAB Technologies BV and Ebben Partners BV, both based in the Netherlands. 

\bibliographystyle{pai2018}
\bibliography{literature}

\begin{thebibliography}{15}
\providecommand{\natexlab}[1]{#1}
\providecommand{\url}[1]{\texttt{#1}}
\expandafter\ifx\csname urlstyle\endcsname\relax
  \providecommand{\doi}[1]{doi: #1}\else
  \providecommand{\doi}{doi: \begingroup \urlstyle{rm}\Url}\fi

\bibitem[Cressey(1951)]{cressey2-1951}
Cressey, D.
\newblock Why do trusted persons commit fraud? a social-psychological study of
  defalcators.
\newblock \emph{Journal of Accountancy}, 92:\penalty0 576, 1951.

\bibitem[Devlin et~al.(2018)Devlin, Chang, Lee, and Toutanova]{Devlin2019}
Devlin, J., Chang, M.-W., Lee, K., and Toutanova, K.
\newblock Bert: Pre-training of deep bidirectional transformers for language
  understanding.
\newblock \emph{arXiv preprint arXiv:1810.04805}, 2018.

\bibitem[Fong et~al.(2008)Fong, Roussinov, and
  Skillicorn]{FongSzeWangandRoussinovDmitriandSkillicorn2008}
Fong, S., Roussinov, D., and Skillicorn, D.~B.
\newblock {Detecting word substitutions in text}.
\newblock In \emph{IEEE Transactions on Knowledge and Data Engineering},
  volume~20, pp.\  1067--1076. IEEE, 2008.

\bibitem[Heinzerling(2019)]{heinzerling2019cleverhans}
Heinzerling, B.
\newblock Nlp's clever hans moment has arrived.
\newblock \emph{The Gradient}, 2019.

\bibitem[Heuer(1999)]{heuer1999psychology}
Heuer, R.~J.
\newblock \emph{Psychology of intelligence analysis}.
\newblock Center for the Study of Intelligence, 1999.

\bibitem[Jabbari et~al.(2008)Jabbari, Allison, and Guthrie]{Jabbari2008}
Jabbari, S., Allison, B., and Guthrie, L.
\newblock {Using a probabilistic model of context to detect word obfuscation}.
\newblock In \emph{Proceedings of the 6th International Conference on Language
  Resources and Evaluation, LREC 2008}, pp.\  2216--2220, 2008.

\bibitem[Kassem \& Higson(2012)Kassem and Higson]{Kassem2012}
Kassem, R. and Higson, A.
\newblock The new fraud triangle model.
\newblock \emph{Journal of emerging trends in economics and management
  sciences}, 3\penalty0 (3):\penalty0 191--195, 2012.

\bibitem[Kingma \& Ba(2014)Kingma and Ba]{Kingma2015}
Kingma, D.~P. and Ba, J.
\newblock Adam: A method for stochastic optimization.
\newblock \emph{arXiv preprint arXiv:1412.6980}, 2014.

\bibitem[Magu \& Luo(2018)Magu and Luo]{Magu2018}
Magu, R. and Luo, J.
\newblock Determining code words in euphemistic hate speech using word
  embedding networks.
\newblock In \emph{Proceedings of the 2nd Workshop on Abusive Language Online
  (ALW2)}, pp.\  93--100, 2018.

\bibitem[Magu et~al.(2017)Magu, Joshi, and Luo]{Magu2017}
Magu, R., Joshi, K., and Luo, J.
\newblock Detecting the hate code on social media.
\newblock \emph{arXiv preprint arXiv:1703.05443}, 2017.

\bibitem[Niven \& Kao(2019)Niven and Kao]{DBLP:journals/corr/abs-1907-07355}
Niven, T. and Kao, H.
\newblock Probing neural network comprehension of natural language arguments.
\newblock \emph{CoRR}, abs/1907.07355, 2019.
\newblock URL \url{http://arxiv.org/abs/1907.07355}.

\bibitem[Paszke et~al.(2019)Paszke, Gross, Massa, Lerer, Bradbury, Chanan,
  Killeen, Lin, Gimelshein, Antiga, et~al.]{paszke2019pytorch}
Paszke, A., Gross, S., Massa, F., Lerer, A., Bradbury, J., Chanan, G., Killeen,
  T., Lin, Z., Gimelshein, N., Antiga, L., et~al.
\newblock Pytorch: An imperative style, high-performance deep learning library.
\newblock In \emph{Advances in neural information processing systems}, pp.\
  8026--8037, 2019.

\bibitem[Peters et~al.(2018)Peters, Neumann, Iyyer, Gardner, Clark, Lee, and
  Zettlemoyer]{peters2018deep}
Peters, M.~E., Neumann, M., Iyyer, M., Gardner, M., Clark, C., Lee, K., and
  Zettlemoyer, L.
\newblock Deep contextualized word representations.
\newblock \emph{arXiv preprint arXiv:1802.05365}, 2018.

\bibitem[Scholtes(2020)]{scholtes2020}
Scholtes, J.~C.
\newblock Text-mining and ediscovery for big-data audits.
\newblock \emph{European Court Auditors Journal. No 1}, pp.\  133--140, 2020.

\bibitem[Scholtes \& Herik(2021)Scholtes and Herik]{scholtesherik2021}
Scholtes, J.~C. and Herik, H. v.~d.
\newblock Big data analytics for ediscovery.
\newblock In \emph{Legal Big Data, Roland Vogl, Eds.} Edgar Publishing, 2021.

\end{thebibliography}

\end{document}